\documentclass[letterpaper, 10 pt, conference]{ieeeconf}  % Comment this line out if you need a4paper

\IEEEoverridecommandlockouts                              % This command is only needed if 
\usepackage{cite}
\usepackage{amsmath,amssymb,amsfonts}
\usepackage{algorithmic}
\usepackage{algorithm}
\usepackage{graphicx}
\usepackage{textcomp}
\usepackage{xcolor}
\usepackage{booktabs}
\usepackage{longtable}
\usepackage{booktabs}   % for better looking tables
\usepackage{multirow}   % if needed for multirow cells
\usepackage{graphicx}   % if inserting images/figures
\usepackage{stfloats}   % useful for two-column floats at bottom of page (optional)
\usepackage{algorithm}

\usepackage{algorithmic}
\usepackage{subcaption}
\usepackage[utf8]{inputenc}
\usepackage{amsmath,amssymb}
\usepackage{background}

\title{\LARGE \bf
InDRiVE: Intrinsic Disagreement-based Reinforcement for Vehicle
Exploration through Curiosity-Driven Generalized World Model}

\author{Feeza Khan Khanzada$^{1}$ and Jaerock Kwon$^{2}$% <-this % stops a space
\thanks{*This work was supported in part by the National Science
Foundation (NSF) under Grant MRI 2214830.}% <-this % stops a space
\thanks{$^{1}$Feeza Khan Khanzada and $^{2}$Jaerock Kwon are with the Department of Electrical and Computer Engineering, University of Michigan-Dearborn,
        4901 Evergreen Rd, Dearborn, MI 48128, United States.
        {\tt\small \{feezakk, jrkwon\}@umich.edu}}% 
}

\begin{document}

\maketitle
\thispagestyle{empty}
\pagestyle{empty}

\backgroundsetup{
  scale=1,
  angle=0,                      % we rotate the text itself, not the entire "background"
  placement=left,               % place near the left edge
  hshift=-10.5cm,
  color=black,
  opacity=1,                    % 1 = fully opaque
  contents={%
    % Rotate the text 90 degrees within a parbox
    \rotatebox{90}{%
      \parbox{23cm}{% Adjust width so the entire text fits on one line
        \small % or \footnotesize, etc.
        This work has been submitted to the IEEE for possible publication. Copyright may be transferred without notice, after which this version may no longer be accessible.
      }%
    }%
  },
}

%%%%%%%%%%%%%%%%%%%%%%%%%%%%%%%%%%%%%%%%%%%%%%%%%%%%%%%%%%%%%%%%%%%%%%%%%%%%%%%%
\begin{abstract}

Model-based Reinforcement Learning (MBRL) has emerged as a promising paradigm for autonomous driving, where data efficiency and robustness are critical. Yet, existing solutions often rely on carefully crafted, task-specific extrinsic rewards, limiting generalization to new tasks or environments. In this paper, we propose InDRiVE (Intrinsic Disagreement-based Reinforcement for Vehicle Exploration), a method that leverages purely intrinsic, disagreement-based rewards within a Dreamer-based MBRL framework. By training an ensemble of world models, the agent actively explores high-uncertainty regions of environments without any task-specific feedback. This approach yields a task-agnostic latent representation, allowing for rapid zero-shot or few-shot fine-tuning on downstream driving tasks such as lane following and collision avoidance. Experimental results in both seen and unseen environments demonstrate that InDRiVE achieves higher success rates and fewer infractions compared to DreamerV2 and DreamerV3 baselines—despite using significantly fewer training steps. Our findings highlight the effectiveness of purely intrinsic exploration for learning robust vehicle control behaviors, paving the way for more scalable and adaptable autonomous driving systems.

% \footnote{This work has been submitted to the IEEE for possible publication. Copyright may be transferred without notice, after which this version may no longer be accessible.}

\end{abstract}

% \SetWatermarkText{%
% This work has been submitted to the IEEE for possible publication.%
% ~Copyright may be transferred without notice, after which this version may no longer be accessible.%
% }

% \SetWatermarkScale{0.05} % Adjust scale as desired
% \SetWatermarkAngle{90}
% \SetWatermarkLightness{0.8}

% % Set horizontal position (xpos): 0 = left edge, 1 = right edge
% \SetWatermarkHorCenter{0.015\paperwidth} % close to the left edge
% % Set vertical position (ypos): 0 = bottom edge, 1 = top edge
% \SetWatermarkVerCenter{0.5\paperheight} % vertically centered

%%%%%%%%%%%%%%%%%%%%%%%%%%%%%%%%%%%%%%%%%%%%%%%%%%%%%%%%%%%%%%%%%%%%%%%%%%%%%%%%
\section{Introduction}

Model-Based Reinforcement Learning (MBRL) has been making significant strides in the robotics domain, offering a compelling alternative to model-free reinforcement learning by focusing on building an internal model of the environment before direct interaction. This approach has shown tremendous potential in reducing training time, creating more generalized models, and mitigating uncertainties. However, the reward-centric nature of traditional reinforcement learning algorithms poses a fundamental challenge to achieving generalization, particularly in scenarios with sparse rewards. Both model-free and MBRL algorithms often rely heavily on task-specific rewards, which limits their adaptability and efficiency in novel environments.

Inspired by neuroscience, curiosity-based learning offers a promising avenue to overcome these limitations \cite{aubret_information-theoretic_2023}. In humans, curiosity drives learning through exploration and the accumulation of experiences, often independent of immediate external rewards \cite{pathak_curiosity-driven_2017}. By seeking novel states in the environment, agents can enhance their exploration capabilities using prediction error from an inverse dynamic model as a measure of novelty. Subsequent studies have refined and expanded on this idea, demonstrating its efficacy in training RL models for better generalization and uncertainty quantification \cite{pathak_self-supervised_2019}\cite{burda_exploration_2018}. Applications of intrinsic motivation have also been explored in domains such as autonomous vehicles, particularly for handling sparse rewards and improving exploration, as highlighted in the related work section.

Despite the progress, there is no comprehensive study that leverages intrinsic motivation to train an MBRL agent for generalization across task-agnostic extrinsic reward functions. Specifically, the ability to train a single agent that can adapt to diverse tasks like lane following and collision avoidance in a zero-shot or few-shot learning setting is largely unexplored. From the prior work surveyed, three major research gaps that capture our attention are:
\begin{itemize}

    \item Although MBRL methods excel in sample efficiency and have been validated on various tasks, they frequently rely on domain or task-specific reward structures and lack evidence of extensive multi-task generalization.
    \item Intrinsic motivation has been shown to improve exploration, but most applications still augment rather than replace extrinsic rewards. There is a lack of studies examining a complete reliance on intrinsic rewards to build a highly adaptable world model.
    \item While some efforts address specific driving tasks, a single agent that can adapt to downstream tasks like lane following and collision avoidance in zero- or few-shot settings remains underexplored.
\end{itemize}

% \begin{figure}[t]
%     \centering
%     \begin{subfigure}{0.3\linewidth}
%         \centering
%         \includegraphics[width=\linewidth]{images/taskpic1.pdf}
%         \caption{}
%         \label{fig:subfig1}
%     \end{subfigure}%
%     \hfill
%     \begin{subfigure}{0.3\linewidth}
%         \centering
%         \includegraphics[width=\linewidth]{images/taskpic2.pdf}
%         \caption{}
%         \label{fig:subfig2}
%     \end{subfigure}
%     \caption{}
%     \label{fig:yourfigure}
% \end{figure}

% \begin{figure}[!t]  % 't' places the figure at the top of the page
%   \centering
%   % Use a smaller height to reduce vertical space; keepaspectratio ensures no distortion.
%   \includegraphics[height=0.135\textheight, keepaspectratio]{images/taskpic.pdf}
%   \caption{Downstream tasks in Autonomous Driving}
%   \label{fig:single_figure}
% \end{figure}

Addressing these gaps, we introduce InDRiVE (Intrinsic Disagreement-based Reinforcement for Vehicle Exploration), which leverages a Dreamer-based MBRL agent. InDRiVE relies solely on ensemble model disagreement for intrinsic motivation, enabling the agent to learn a robust, task-agnostic latent world model. Our objective is to facilitate zero-shot or few-shot fine-tuning across diverse driving tasks, thus minimizing training time and reducing reliance on manual reward engineering for real-world deployment. Following is the list of contributions of InDRiVE through this research:

\begin{itemize} 

    \item To the best of our knowledge, InDRiVE is the first study to train an ego-vehicle \emph{exclusively} with intrinsic rewards, leveraging latent disagreement among an ensemble of world models (based on the \cite{sekar_planning_2020}). This eliminates the need for hand-crafted task rewards, relying solely on uncertainty-based signals to build a robust, task-agnostic representation of the environment. 
    
    \item The resulting world model supports zero-shot and few-shot adaptation to real driving tasks (e.g., lane-following, collision avoidance), drastically reducing domain-specific reward engineering.
    
    \item Through purely intrinsic exploration, our approach yields a versatile world model capable of zero-shot and few-shot transfer to downstream driving tasks such as lane-following and collision avoidance. This demonstrates that the learned model is not only comprehensive but also quickly adaptable to practical driving objectives, significantly reducing the need for domain-specific reward engineering.

    % \item We empirically compare the performance of our intrinsically-trained Dreamer-based agent against both DreamerV2 and DreamerV3 baselines equipped with traditional (extrinsic) task rewards. Our experiments measure success rate, infraction rate, and learning efficiency, shedding light on the robustness and real-world relevance of a fully self-supervised exploration policy. 
    
    \item Our findings confirm that fully intrinsic reward mechanisms are both viable and beneficial for high-dimensional, safety-critical domains like autonomous driving, paving the way for broader self-supervised MBRL solutions.
    
\end{itemize}

By capitalizing on intrinsic model disagreement signals, InDRiVE achieves robust exploration, rapid adaptation to new tasks, and a streamlined reward design pipeline—pointing toward more scalable, self-supervised solutions for future autonomous vehicles.

\section{Related Work}

MBRL has transitioned from a theoretical construct to a practical solution for autonomous vehicle (AV) control, driven by advances in model fidelity, planning algorithms, and deep neural networks \cite{ha_recurrent_2018}. Unlike model-free methods, which rely primarily on trial-and-error, MBRL incorporates a learned world model of the environment to enable look-ahead planning and improve data efficiency. In the context of autonomous driving, these learned models can anticipate future states and rewards, allowing for safer decision-making and reduced real-world experimentation \cite{gao_dream_2024}\cite{hu_model-based_2022}. Recent work in simulation platforms such as CARLA \cite{Dosovitskiy17} has demonstrated that world-model-based planners can imagine a diverse range of upcoming scenarios before executing actions, thus mitigating safety risks and addressing data scarcity by synthesizing additional training samples \cite{kiran_deep_2022}. Continued innovations like latent state abstraction, uncertainty-aware modeling, and online adaptation further reduce the gap between purely simulated training and real-world deployment \cite{hafner_dream_2020}\cite{hafner_mastering_2022}.  Moreover, while most prior efforts focus on on-road driving scenarios, recent analytical study on off-road autonomy found that selecting the right image region-of-interest and using a larger training dataset significantly improves the performance of vision-based end-to-end lateral control \cite{khanzada2024analytical}. Such findings highlight the importance of data representation and collection strategies, which could similarly benefit model-based methods by ensuring that learned representations capture critical environmental cues across diverse driving conditions.

Intrinsic Motivation (IM) and curiosity-driven exploration have emerged as essential mechanisms for guiding agents in sparse-reward or high-dimensional environments, where extrinsic feedback is rare or too costly to define \cite{pathak_curiosity-driven_2017}\cite{burda_large-scale_2018-1}. IM provides agents with self-generated reward signals that encourage exploration, often by rewarding novelty, uncertainty, or prediction error \cite{meyer_possibility_1991}\cite{stadie_incentivizing_2015}. Notable curiosity-based approaches include the Intrinsic Curiosity Module (ICM) \cite{pathak_curiosity-driven_2017} and Random Network Distillation (RND) \cite{burda_exploration_2018}, both of which incentivize agents to visit unfamiliar or surprising states. Such methods have been successfully applied to robotic systems and video game domains, enabling agents to learn skills in the absence of dense external rewards \cite{burda_large-scale_2018-1}\cite{oudeyer_intrinsic_2007}. However, purely intrinsic exploration can lead agents to fixate on irrelevant or noisy events, spurring interest in techniques that combine curiosity with additional constraints or memory mechanisms to ensure meaningful, goal-relevant exploration \cite{raileanu_ride_2020-1}.

In the realm of autonomous driving, prior research has mostly leveraged intrinsic motivation as a complementary signal rather than a primary training objective \cite{noauthor_imagine-2-drive_nodate}\cite{gao_dream_2024}\cite{hu_model-based_2022}. Typical reinforcement learning frameworks for driving rely on task-specific reward functions (e.g., measuring route progress, penalizing collisions, or encouraging lane-keeping) \cite{toromanoff_end--end_2020}, often augmented with a small curiosity bonus to expedite convergence. While this hybrid approach can alleviate some exploration hurdles, it still anchors the learned policy to a particular extrinsic objective, reducing its flexibility to generalize across tasks or conditions. . Additionally, exploration in autonomous driving requires careful consideration of safety and real-world feasibility; purely random or naive exploration is not viable in practice, further complicating the application of intrinsic rewards \cite{codevilla_exploring_2019}.

Crucially, existing literature lacks studies that train a full end-to-end driving policy \emph{exclusively} via intrinsic rewards. While purely curiosity-driven methods have been demonstrated in simpler continuous-control scenarios \cite{sekar_planning_2020}\cite{burda_large-scale_2018}, no prior work has shown that an autonomous vehicle agent can acquire complex driving behaviors (such as collision avoidance or lane-following) without relying on explicit, task-specific feedback. This gap is particularly significant given the potential advantages of a fully task-agnostic paradigm in which the agent discovers relevant driving skills independently and subsequently fine-tunes to specific tasks with minimal overhead. Our work aims to bridge this gap by integrating an ensemble-based \emph{model disagreement} signal, inspired by the \cite{sekar_planning_2020}, into a Dreamer-based agent \cite{hafner_dream_2020}, allowing the vehicle to learn a robust world model in CARLA solely through intrinsic exploration signals. Ultimately, this approach seeks to demonstrate that internal disagreement metrics can serve as a standalone training driver, paving the way for efficient, flexible, and generalized autonomous driving policies.

A recent analytical study on off-road autonomy found that selecting the right image region-of-interest and using a larger training dataset significantly improves the performance of vision-based end-to-end lateral control \cite{khanzada2024analytical}

\section{Methodology}

% \begin{figure*}[t]
%     \centering
% %    \includegraphics[width=0.9\textwidth]{images/Methodology.pdf}
%     \includegraphics[height=5cm]{images/Methodology.pdf}
%     \caption{Overview of the proposed InDRiVE architecture combining an intrinsic reward signal 
%     with an MBRL framework. The top section shows an 
%     encoder transforming raw images into latent states $s_t$, which are then passed 
%     to a policy $\pi_{\theta}$ to output actions $a_t$. The policy also receives an 
%     intrinsic reward (\textit{IR}), computed via the variance of multiple future-state 
%     predictions $\{h^k_{t+1}\}$ in the ensemble (shown in the green box). At each time 
%     step, the agent updates its latent state $s_{t+1}$ and value function $V_{t+1}$, 
%     using the intrinsic reward signal $r^i_{t+1}$ to encourage exploration. This 
%     variance-based disagreement among ensemble heads provides curiosity-driven 
%     feedback, guiding the policy to learn robust behaviors without extrinsic reward.}
%     \label{fig:method_overview}
% \end{figure*}

\begin{figure*}[t]
  \centering
  % -- First subfigure --
  \begin{subfigure}[t]{0.69\textwidth} % Larger box
    \centering
    \includegraphics[scale=0.5]{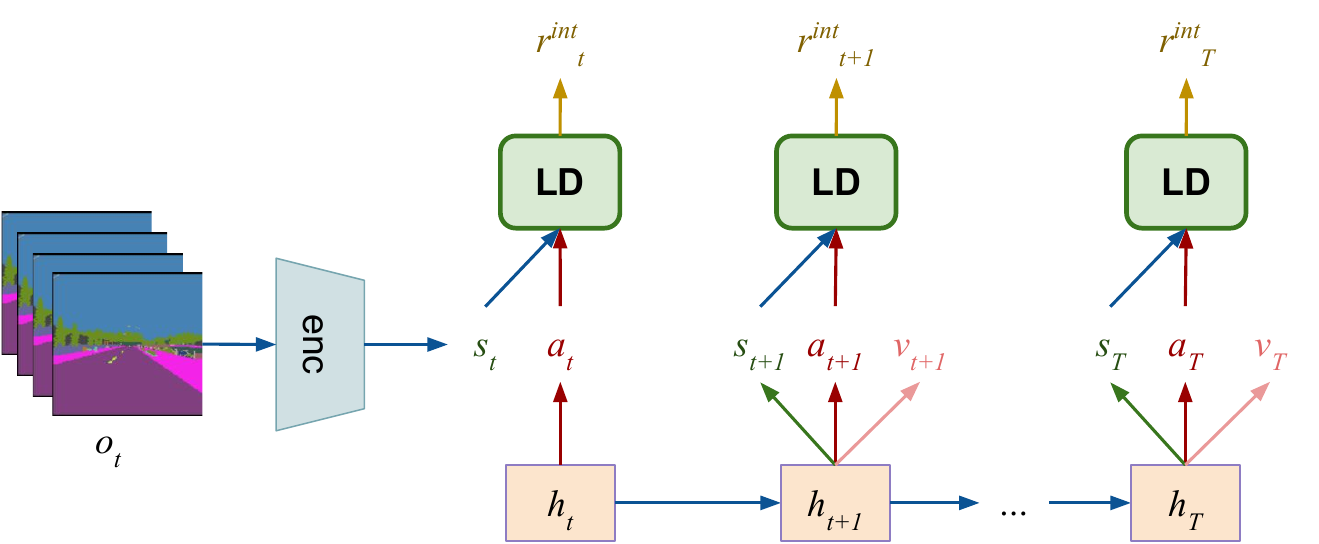}
    \caption{Overview of the InDRiVE Actor Critic Policy.}
    \label{fig:method_overview_a}
  \end{subfigure}
  \hfill
  % -- Second subfigure --
  \begin{subfigure}[t]{0.3\textwidth} % Smaller box
    \centering
    \includegraphics[scale=0.45]{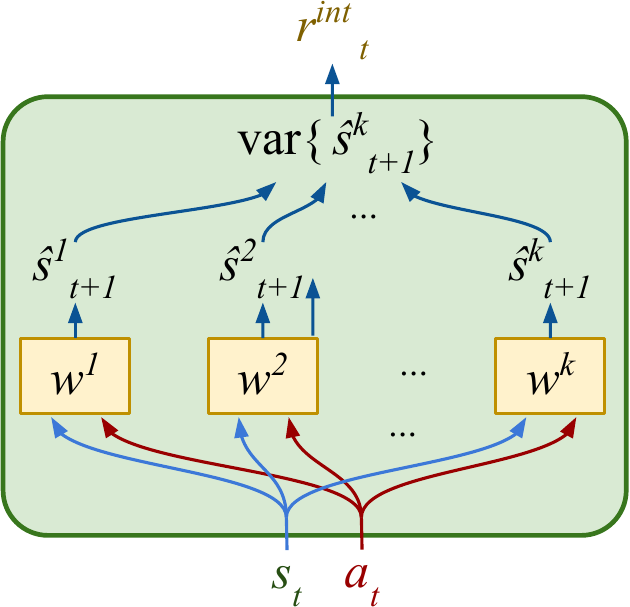}
    \caption{Latent Disagreement (LD) Reward}
    \label{fig:method_overview_b}
  \end{subfigure}

  \caption{%
    Overview of the InDRiVE. 
    (a) An actor critic policy architecture incorporating latent disagreement for exploration. LD is Latent Disagreement in (b). Raw images are encoded into a stochastic latent $s_t$, which is combined with deterministic hidden state $h_t$ to maintain temporal context. The actor--critic policy then outputs an action $a_t$ based on $[s_t, h_t]$. 
    (b) An \textit{ensemble} of forward 
    models predicts potential next states $\hat{s}_{t+1}^{\,k}$ for the same $(s_t, a_t)$. The variance among these predictions yields a latent-disagreement (intrinsic) reward, which, encourages the policy to explore.}
  \label{fig:method_overview}
\end{figure*}

\noindent In this section, we detail our InDRiVE approach, which extends DreamerV3 with an ensemble-based intrinsic exploration mechanism inspired by \cite{sekar_planning_2020} The goal is to train a robust, task-agnostic world model via curiosity-driven exploration, then fine-tune the learned policy with minimal additional effort for specific driving tasks in CARLA. Fig.~\ref{fig:method_overview} presents a high-level overview of InDRiVE alongwith the latent disagreement mechanism.

\subsection{Intrinsic Motivation and World Model}
InDRiVE is an MBRL framework designed for autonomous driving. It adopts the DreamerV3 architecture \cite{hafner_mastering_2024} for its latent world model and planning capabilities while leveraging ensemble disagreement to generate purely intrinsic rewards during an initial exploration phase. This approach is motivated by \cite{sekar_planning_2020}, which demonstrated that self-supervised exploration improves sample efficiency and task generalization. In InDRiVE, we first train the agent solely with intrinsic rewards (no task-specific feedback), yielding a broad coverage of driving scenarios and a capable latent world model. Subsequently, we introduce extrinsic rewards to fine-tune the policy for tasks such as lane following or collision avoidance.

%\subsubsection{Reinforcement Learning and World Model}
We formulate autonomous driving as a Markov Decision Process (MDP) $\mathcal{M} = (\mathcal{S}, \mathcal{A}, p, r, \gamma)$. States $s \in \mathcal{S}$ encapsulate sensor observations, while actions $a \in \mathcal{A}$ correspond to vehicular control inputs (steering, throttle, braking). The transition model $p(s_{t+1} \mid s_t, a_t)$ governs environment dynamics, and $r(s_t, a_t)$ provides task-dependent feedback. In our setting, \emph{intrinsic exploration} replaces task-specific rewards during the initial training phase:
\begin{equation*}
r_t^{\text{int}} \;=\; \text{Disagreement-based curiosity signal},
\end{equation*}
whereas the \emph{fine-tuning} phase introduces extrinsic signals:

\begin{equation*}
r_t^{\text{ext}} \;=\; \text{task-specific reward signal}.
\end{equation*}

We can also combine the extrinsic and intrinsic rewards, where intrinsic reward can be used to augment the reward based training:

\begin{equation}
    r_t \;=\; \alpha\,r_t^{\text{ext}} \;+\; (1-\alpha)\,r_t^{\text{int}},
    \label{eq:combined-reward}
\end{equation}
with $\alpha \in [0,1]$ controlling the weighting between extrinsic and intrinsic rewards.

We adopt a Recurrent State-Space Model (RSSM) to learn a compact representation of high-dimensional sensory inputs (e.g., images) and predict future observations and rewards. Following the Dreamer framework~\cite{hafner_mastering_2024}, the RSSM consists of four main components:

\begin{itemize}
    \item Encoder $q_{\phi}(z_t \mid s_t)$: Converts raw observations $s_t$ into a stochastic latent state $z_t$.
    \item Recurrent Core (GRU): Maintains a hidden state $h_t$, summarizing past latent states and actions.
    \item Transition Model $p_{\phi}(z_{t+1} \mid z_t, a_t, h_t)$: Predicts the next latent state $z_{t+1}$ given the current latent state, action $a_t$, and hidden state $h_t$.
    \item Decoder $p_{\phi}(s_t \mid z_t)$: Reconstructs or \textit{imagines} the original observation $s_t$ from the latent state $z_t$.
\end{itemize}

Additionally, we include a reward predictor $p_{\phi}(r_t \mid z_t, h_t)$ to model the immediate reward, and a discount (or continuation) predictor $p_{\phi}(\gamma_t \mid z_t, h_t)$ to handle episode termination. At each time step, we thus have:
\begin{align}
    h_t &= \text{GRU}(h_{t-1}, z_{t-1}, a_{t-1}), \\
    z_t &\sim q_{\phi}(z_t \mid s_t, h_t), 
\end{align}
with the transition prior
\begin{equation}
    p_{\phi}(z_{t+1} \mid z_t, a_t, h_{t+1}).
\end{equation}

We jointly optimize the encoder, decoder, transition, reward, and discount networks. The training loss, inspired by the variational Evidence Lower Bound (ELBO), can be expressed as:
\begin{equation}
\label{eq:rssm_objective}
\begin{split}
    \mathcal{L}_\text{model}(\phi)
    \;=\;& \mathbb{E}_{q_{\phi}}\Bigl[
          -\ln\, p_{\phi}(s_t \mid z_t)
          \;-\;\ln\, p_{\phi}(r_t \mid z_t, h_t)
          \Bigr] \\
    &\quad +\;\beta \,\mathbb{E}_{q_{\phi}}\!\Bigl[
        D_{\mathrm{KL}}\!\bigl(q_{\phi}(z_t \mid s_t, h_t)\;\|\;p_{\phi}(z_t \mid h_t)\bigr)
    \Bigr]\\
    &\quad +\;\lambda_{\gamma}\,\mathbb{E}_{q_{\phi}}\!\Bigl[
        -\ln\,p_{\phi}(\gamma_t \mid z_t, h_t)
    \Bigr],
\end{split}
\end{equation}
where:
\begin{itemize}
    \item $-\ln\, p_{\phi}(s_t \mid z_t)$ is the \emph{reconstruction loss}, penalizing the model for poor observation predictions.
    \item $-\ln\, p_{\phi}(r_t \mid z_t, h_t)$ is the \emph{reward prediction loss}.
    \item $D_{\mathrm{KL}}\!\bigl(q_{\phi}(z_t \mid s_t, h_t)\;\|\;p_{\phi}(z_t \mid h_t)\bigr)$ is the \emph{KL divergence} between the encoder posterior and the transition prior, encouraging compact and consistent latent states.
    \item $\beta$ scales or clips the KL term (the \emph{free-bits} heuristic~\cite{hafner_mastering_2022}) so that the model retains sufficient representational capacity without collapsing.
    \item $-\ln\,p_{\phi}(\gamma_t \mid z_t, h_t)$ is an optional \emph{discount/continuation loss} (weighted by $\lambda_{\gamma}$) that helps the model account for terminal states.
\end{itemize}

\noindent We sample short latent-rollout sequences from a replay buffer of past trajectories, optimize \eqref{eq:rssm_objective} in mini-batches, and update the model parameters $\phi$ via stochastic gradient descent.

\noindent Once trained, the RSSM provides a forward model for \emph{imagined rollouts}: starting from a real or latent-encoded state, the model predicts future states, rewards, and discounts, thereby enabling policy learning and planning entirely within the compact latent space.

\subsection{Ensemble Disagreement for Intrinsic Exploration}
\label{sec:ensemble-disagreement}
We incorporate \emph{ensemble disagreement} to drive curiosity, building on the self-supervised exploration scheme introduced by \cite{pathak_self-supervised_2019}. Specifically, we train $K$ lightweight forward dynamics models, each predicting the next latent state $s_{t+1}$ given $(s_t, a_t)$. Let $\mu_k(s_t, a_t)$ denote the prediction of the $k$-th model. The intrinsic reward $r_t^{\text{int}}$ is computed as the variance of these predictions:
\begin{equation}
\label{eq:disagreement}
    r_t^{\text{int}} \;=\; 
    \mathrm{Var}\Bigl(\mu_1(s_t,a_t),\,\mu_2(s_t,a_t),\dots,\mu_K(s_t,a_t)\Bigr).
\end{equation}
High disagreement indicates unexplored or uncertain regions, incentivizing the policy to gather data where the world model is less confident. As training progresses, this promotes coverage of diverse states and reduces model uncertainty in safety-critical scenarios.

\subsection{Steering Loss Function}
\label{sec:steering-loss}

To encourage smooth driving behavior, we introduce a steering loss function inspired by \cite{li_think2drive_2024}, adapted to penalize excessively large steering angles. Let $a_t^{(\text{steer})}$ denote the steering command at time $t$, measured in the range $[-1, 1]$ (left to right turn). We define:
\begin{equation}
    r_{\text{steer}}(a_t) \;=\;
    \begin{cases}
        -\lambda, & \text{if } \bigl\lvert a_t^{(\text{steer})}\bigr\rvert \;>\;\delta, \\
        0,        & \text{otherwise},
    \end{cases}
    \label{eqn:steer-penalty}
\end{equation}
where $\lambda > 0$ and $\delta \in (0,1)$ is a steering-angle threshold. In practice, we set $\lambda = 0.5$ and incorporate this penalty term during training. This additional cost biases the policy to avoid extreme steering angles, thereby promoting smoother, more stable navigation without preventing necessary turns.

\subsection{Training Procedure}
\label{sec:training-procedure}
Algorithm~\ref{alg:indrive-training} summarizes the two-phase training pipeline:

\begin{algorithm}[t]
\caption{\textbf{InDRiVE Training Procedure}}
\label{alg:indrive-training}
\begin{algorithmic}[1]
\REQUIRE Environment $\mathcal{E}$ (CARLA), replay buffer $\mathcal{D}$, number of ensemble models $K$, exploration steps $N_{\text{explore}}$, fine-tuning steps $N_{\text{fine}}$.
\STATE Initialize parameters of DreamerV3 world model $\phi$, policy $\pi_\theta$, value network $v_\theta$, and ensemble models $\{\mu_k\}_{k=1..K}$.
\STATE $\mathcal{D} \leftarrow \{\}$ \quad (empty replay buffer)

\FOR{step = 1 \textbf{ to } $N_{\text{explore}}$}
    \STATE Roll out policy $\pi_\theta$ in $\mathcal{E}$ for $T$ steps to collect $\{(o_t, a_t, o_{t+1})\}$.
    \STATE Encode $s_t \leftarrow q_\phi(s_t \mid o_t)$.
    \STATE Compute disagreement $r_t^{\text{int}}$ \;\;via \;Eq.~(\ref{eq:disagreement}).
    \STATE $\mathcal{D} \leftarrow \mathcal{D} \cup \{(s_t,a_t,r_t^{\text{int}},s_{t+1})\}$.
    \STATE Update DreamerV3 world model \& ensemble models using ELBO-based loss (Eq.~(\ref{eq:rssm_objective})).
    \STATE Update $\pi_\theta$, $v_\theta$ via imagination in the latent space, optimizing intrinsic returns.
\ENDFOR

\STATE \COMMENT{Fine-Tuning for Task-Specific Rewards}
\FOR{$\text{step} = 1$ \textbf{ to } $N_{\text{fine}}$}
    \STATE Introduce extrinsic reward $r_{\text{ext}}$ for downstream task (e.g., collision avoidance).
    \STATE $r_t = r_t^{\text{ext}} $
    \STATE \textit{(zero-shot): no additional data collection.}
    \STATE \textit{(few-shot): gather limited on-policy data to refine } $\phi, \theta$.
\ENDFOR

\RETURN Optimized policy $\pi_\theta$ and world model parameters $\phi$.

\end{algorithmic}
\end{algorithm}

% \begin{algorithm}[t]
% \caption{\textbf{InDRiVE Training Procedure}}
% \label{alg:indrive-training}
% \begin{algorithmic}[1]
    
%     \STATE Initialize model parameters
%     \STATE Train model on dataset
%     \STATE Evaluate performance
% \end{algorithmic}
% \end{algorithm}

\noindent
\textbf{Phase 1: Task-Agnostic Exploration:} The agent explores the CARLA environment by maximizing the ensemble-disagreement reward, augmented with the steering penalty for stable control. This phase yields a broad coverage of driving states and a well-trained world model without relying on task-specific guidance.

\noindent
\textbf{Phase 2: Task-Specific Fine-Tuning:} We then introduce the extrinsic driving objective (lane following and collision avoidance). The policy learns to balance this task reward with the residual intrinsic signal and steering loss. In many cases, zero-shot adaptation is possible, as the agent’s learned representation already encodes crucial driving behaviors. Otherwise, a small number of additional training episodes is sufficient for few-shot adaptation, drastically reducing total sample complexity compared to purely extrinsic-driven training.

\noindent Overall, this two-phase approach demonstrates how self-supervised exploration can bootstrap a robust world model, leading to faster and more versatile task adaptation in autonomous driving. We use the CARLA simulator as our primary testbed, taking advantage of its:
\begin{itemize}
    \item \textit{Realistic sensor data}: RGB camera, LiDAR, GPS, and odometry information,
    \item \textit{Complex traffic scenarios}: dynamic vehicles, pedestrians, traffic lights, and multi-lane roads,
    \item \textit{Configurable weather and lighting conditions}: enabling diverse scenarios for robust exploration.
\end{itemize}

\section{Experimental Setup}
\label{sec:experimental-setup}

This section details the experimental framework for assessing our proposed approach. We begin by introducing the CARLA simulation environment and the tasks under consideration, followed by the two-phase training procedure. We then describe the baseline methods, hyperparameter configurations, and the metrics used for evaluation.

\subsection{Environment Setup}
\label{sec:env-setup}

We focus on two CARLA towns \texttt{Town01} (a small town with a river and bridges) and \texttt{Town02} (a small town with a mixture of residential and commercial buildings) with moderate traffic density. At each time step, the agent receives a $128 \times 128$ semantic segmentation image, along with throttle and steering angle information. To capture temporal dependencies, we stack four consecutive semantic segmentation frames as a single observation input to the encoder. The agent outputs continuous control commands (steering, throttle, brake).

\subsection{Tasks and Scenarios}
\label{sec:tasks-scenarios}

We consider two representative driving tasks to demonstrate zero-shot and few-shot performance:

\begin{itemize}
    \item Lane Following (LF): The agent must maintain its lane position while traveling at a safe speed.
    \item Collision Avoidance (CA): The agent must avoid colliding with other vehicles and obstacles in real-time traffic scenarios.
\end{itemize}

% \subsubsection{Terminal Conditions}
% \label{sec:termination-conditions}

Episodes terminate upon any of the following events:
\begin{itemize}
    \item Collision: The agent collides with another vehicle, pedestrian, or static obstacle.
    \item Wrong Direction: The agent drives in the opposite direction of the intended lane.
    \item Off-Road Driving: The agent leaves the drivable area.
    \item Vehicle Stall: The agent's velocity falls below a minimal threshold (e.g., 1~km/h) for an extended period (e.g., 1~minute).
    \item Episode Completion: The agent successfully complete the number of steps assigned for the episode without any lane violation and collisions. 
\end{itemize}

\subsection{Two-Phase Training Procedure}
\label{sec:training-phases}

The learning process is divided into two main phases: (i) \textit{intrinsic exploration} for building a general-purpose world model, and (ii) \textit{task-specific fine-tuning} that leverages this model for downstream tasks.

\subsubsection{Task-Agnostic Exploration} \label{sec:intrinsic}

During Phase~1, we train a task-agnostic InDRiVE solely using an intrinsic reward derived from ensemble disagreement. Specifically, we randomize key environment parameters (e.g., weather, traffic density) every 10,000 steps to ensure diverse experiences, then roll out the current policy for 1,000 steps and store all transitions in a replay buffer. Afterward, we update the ensemble of forward dynamics models, along with the encoder-decoder modules and policy/value networks, in latent space using the intrinsic reward $r_t^{\text{int}}$. This cycle of randomization, data collection, and model updating is repeated until a predetermined maximum of environment interactions (e.g., 50K steps) is reached.

\subsubsection{Task-Specific Fine-Tuning} \label{sec:fine-tuning}

Following intrinsic exploration, we evaluate and refine the agent’s performance on downstream tasks (e.g., lane following or collision avoidance) through both zero-shot and few-shot evaluations. For zero-shot evaluation, we freeze the world model parameters (encoder, decoder, and ensemble) and directly test the policy without further training, recording success rates and infractions to gauge initial performance. For few-shots evaluation, we introduce a task-specific extrinsic reward $r_t^{\text{extr}}$, collect a small batch of new data, and update the policy and value networks by applying extrinsic rewards. We then measure the resultant performance gains to assess how effectively the agent adapts to the target task.

% \begin{figure*}[t] 
%   \centering

%   \begin{subfigure}[t]{0.32\textwidth}
%     \centering
%     \includegraphics[width=\linewidth,
%       trim={0.3in 0.2in 0.3in 0.2in},clip]
%       {images/lf.pdf}
%     \caption{}
%   \end{subfigure}
%   \hfill
%   \begin{subfigure}[t]{0.32\textwidth}
%     \centering
%     \includegraphics[width=\linewidth,
%       trim={0.3in 0.2in 0.3in 0.2in},clip]
%       {images/ca.pdf}
%     \caption{}
%   \end{subfigure}
%   \hfill
%   \begin{subfigure}[t]{0.32\textwidth}
%     \centering
%     \includegraphics[width=\linewidth,
%       trim={0.3in 0.2in 0.3in 0.2in},clip]
%       {images/lc.pdf}
%     \caption{}
    
%   \end{subfigure}

%   \caption{%
%    Average reward rates of InDRiVE (red), DreamerV3 (blue), and DreamerV2 (green) across three CARLA driving tasks: (a) Lane Following, (b) Collision Avoidance, and (c) Lane Following combined with Collision Avoidance. In each subfigure, the y‐axis denotes the reward rate while the x‐axis represents the number of environment steps. The gray area after 500K steps indicates the start of InDRiVE’s finetuning phase (few‐shot learning). Despite being trained on the extrinsic reward for fewer steps (10K), InDRiVE (red) converges to near‐optimal performance in all three tasks—surpassing both Dreamer baselines—and demonstrates superior sample efficiency and training stability overall.
%   }
%   \label{fig:reward_rate}
%   \label{fig:method_overview}
% \end{figure*}

\begin{figure*}[t]
  \centering
  
  % Subfigure (a)
  \begin{subfigure}[t]{0.32\textwidth}
    \centering
    \includegraphics[%
      width=\linewidth,
      trim={0.1in 0.1in 0.1in 0.1in},
      clip]{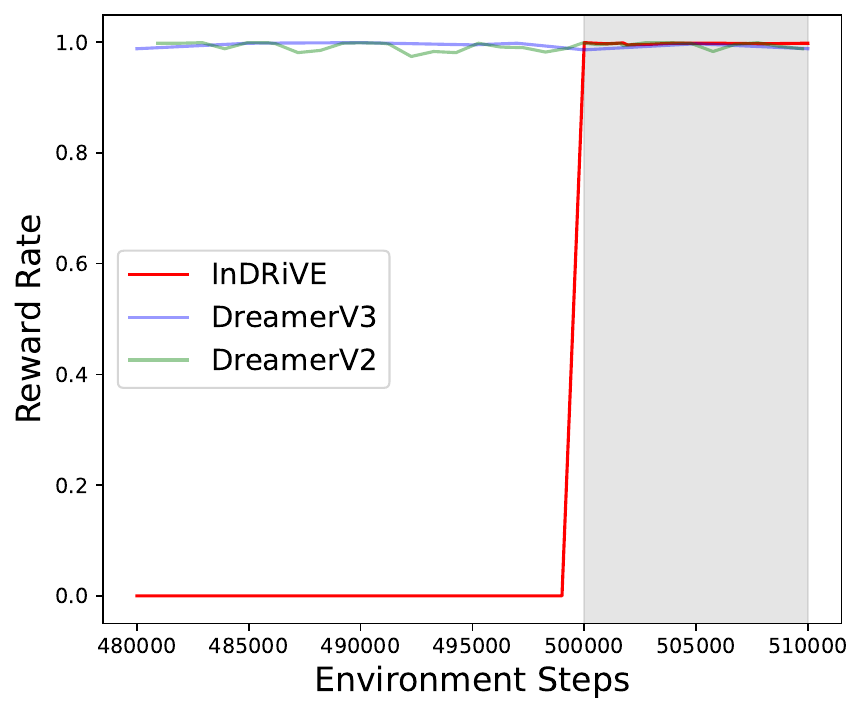}
    \caption{Lane Following}
  \end{subfigure}
  \hfill
  % Subfigure (b)
  \begin{subfigure}[t]{0.32\textwidth}
    \centering
    \includegraphics[%
      width=\linewidth,
      trim={0.1in 0.1in 0.1in 0.1in},
      clip]{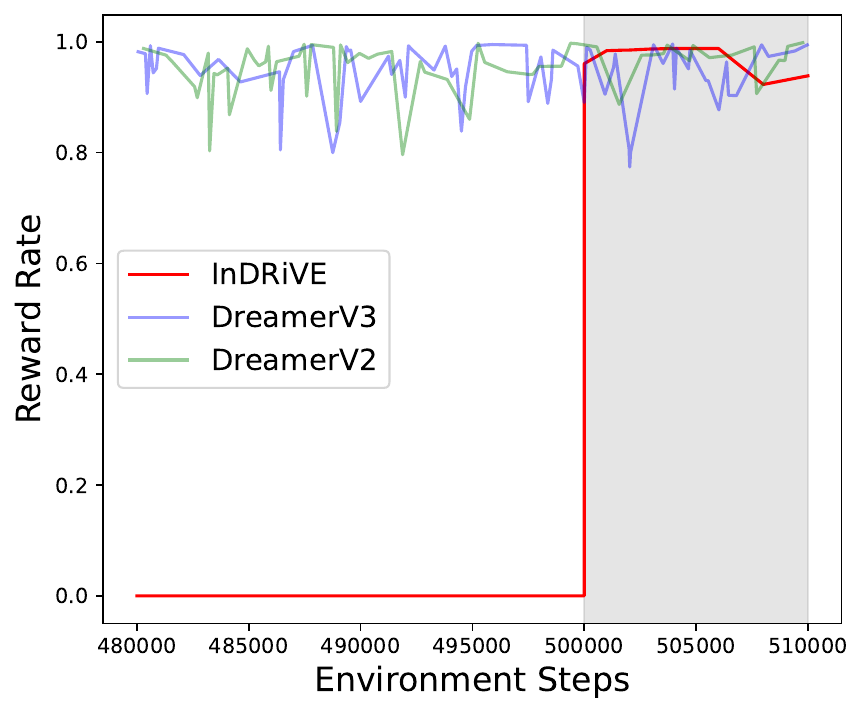}
    \caption{Collision Avoidance}
  \end{subfigure}
  \hfill
  % Subfigure (c)
  \begin{subfigure}[t]{0.32\textwidth}
    \centering
    \includegraphics[%
      width=\linewidth,
      trim={0.1in 0.1in 0.1in 0.1in},
      clip]{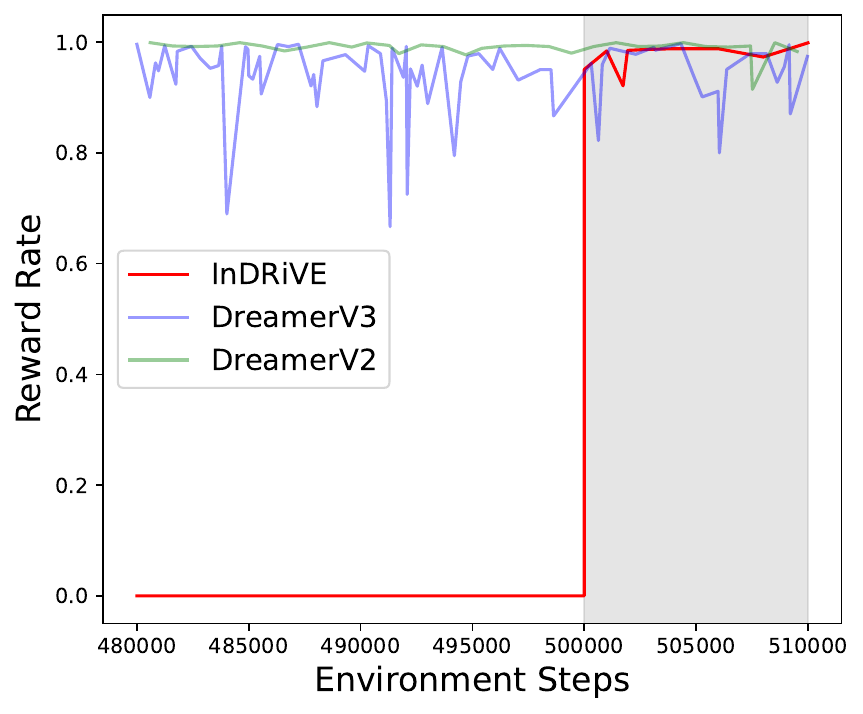}
    \caption{Lane Following + Collision Avoidance}
  \end{subfigure}

  \caption{%
    Average reward rates of InDRiVE (red), DreamerV3 (blue),
    and DreamerV2 (green) across three CARLA driving tasks.
    % (a) Lane Following, (b) Collision Avoidance, and
    % (c) Lane Following combined with Collision Avoidance.
    The gray area after 500K steps indicates the start of
    InDRiVE’s finetuning phase (few‐shot learning). Despite
    being trained on the extrinsic reward for fewer steps
    (10K), InDRiVE (red) converges to near‐optimal performance
    in all three tasks—surpassing both Dreamer baselines—and
    demonstrates superior sample efficiency and training
    stability overall.
  }
  \label{fig:reward_rate}
\end{figure*}

\subsection{Baseline Methods} \label{sec:baselines}

For comparative evaluation, we focus on DreamerV2 (Task-Specific) and DreamerV3 (Task-Specific), both of which train a world model and policy from scratch using only task-specific rewards (e.g., for lane following or collision avoidance), without incorporating any intrinsic rewards. These baselines thus provide a performance and sample-efficiency benchmark for traditional, task-centric learning approaches, enabling a clear assessment of the benefits gained by integrating intrinsic exploration in our method.

Table~\ref{tab:hyperparams} summarizes the key hyperparameters used during the intrinsic exploration phase and the few-shot fine-tuning phase.

\begin{table}[htbp]
    \centering
    \caption{Key Hyperparameters for InDRiVE and Fine-Tuning}
    \label{tab:hyperparams}
    \resizebox{\columnwidth}{!}{%
    \begin{tabular}{lcc}
    \toprule
    \textbf{Hyperparameter} & \textbf{Intrinsic Exploration} & \textbf{Fine-Tuning} \\
    \midrule
    Learning rate (world model)   & $1\times10^{-4}$ & $1\times10^{-4}$ \\
    Learning rate (policy/value)  & $1\times10^{-4}$ & $1\times10^{-4}$ \\
    Ensemble size ($K$)           & 8                & 8 (frozen) \\
    Batch size                    & 64               & 64 \\
    Replay buffer size            & $10^{5}$         & $10^{5}$ \\
    Discount factor ($\gamma$)    & 0.99             & 0.99 \\
    Intrinsic reward weight ($\alpha$) & 1.0 & 0.0 \\
    Training steps                & 50k env steps  & 10K env steps \\
    \bottomrule
    \end{tabular}%
    }
\end{table}

\subsection{Evaluation Metrics}
\label{sec:metrics}

We benchmark InDRiVE on multiple scenarios in the CARLA simulator. Key metrics include:
\begin{itemize}
    \item Success Rate (SR): Rate of successful completion of episode with any lane violation and collision.
    \item Infraction Rate (IR): Rate of rule violations (collisions, lane departures) per episode.
    \item Zero-Shot/Few-Shot Adaptation: Evaluates how well the agent performs the task with no (zero-shot) and minimal (few-shot) additional interactions, highlighting the benefit of curiosity-driven exploration.
\end{itemize}

\noindent
By measuring performance across different tasks, towns, and training regimes, we obtain a comprehensive view of zero-shot and few-shot generalization in complex urban driving scenarios.

\section{Results}

\subsection{Zero-Shot Evaluation}

Table \ref{tab:dreamerv3_results_sr_ir} reports the zero-shot evaluation of DreamerV3, trained in \texttt{Town01} using only a latent disagreement–based intrinsic reward signal when tested in both \texttt{Town01} and \texttt{Town02}. The model is trained for 500K exploration steps, and performance is measured over 50K evaluation steps in each town.

Overall, these results indicate that using InDRiVE in the training phase can yield an agent capable of generalizing from one town to another. While the transfer performance remains slightly lower than the results observed in the training environment, the similarity in success and collision rates suggests that the agent’s learned exploration strategy maintains a degree of robustness across different environments.

% \begin{table}[ht]
%     \centering
%     \caption{Evaluation of DreamerV3(trained in Town01 with only latent disagreement–based reward) on Town01 and Town02. (Zero-Shot Learning)}
%     \label{tab:dreamerv3_results}
%     \resizebox{\columnwidth}{!}{ % Resize table to fit within a column
%     \begin{tabular}{cccccccc}
%         \toprule
%         \textbf{Train} & \textbf{Explore} & \textbf{Eval} & \textbf{Eval} & \textbf{Success} & \textbf{Collisions} & \textbf{Off} & \textbf{Total} \\
%         \textbf{Town} & \textbf{Steps} & \textbf{Town} & \textbf{Steps} & \textbf{Rate (\%)} & \textbf{Count} & \textbf{Road} & \textbf{Eps} \\
%         \midrule
%         Town01 & 500K & Town01 & 50K & 40 & 10 & 12 & 62 \\
%         Town01 & 500K & Town02 & 50K & 41 & 10 & 13 & 64 \\
%         \bottomrule
%     \end{tabular}
%     }
% \end{table}

\begin{table}[ht]
    \centering
    \caption{Zero-Shot Learning Evaluation of InDRiVE on Town01 \& 02}
    \label{tab:dreamerv3_results_sr_ir}
    \resizebox{\columnwidth}{!}{%
    \begin{tabular}{c c | c c | c c}
    \toprule
    %---------------- Header Row 1 ----------------
    \multirow{2}{*}{\textbf{Train Steps}} & \multirow{2}{*}{\textbf{Eval Steps}} 
    & \multicolumn{2}{c|}{\textbf{Eval Town01} (seen)} 
    & \multicolumn{2}{c}{\textbf{Eval Town02} (unseen)} \\
    
    %---------------- Header Row 2 ----------------
    \cmidrule(lr){3-4} \cmidrule(lr){5-6}
     &  
    & \textbf{SR (\%) \textbf{↑}} & \textbf{IR (\%)  \textbf{↓}} 
    & \textbf{SR (\%) \textbf{↑}} & \textbf{IR (\%)  \textbf{↓}} \\
    \midrule

    %---------------- Data Row ----------------
    500K & 50K & 64.52 & 35.48 & 64.06 & 35.94 \\

    \bottomrule
    \end{tabular}%
    }
\end{table}

% \begin{table}[ht]
%     \centering
%     \caption{Zero-Shot Learning Evaluation of InDRiVE on Town01 \& 02}
%     \label{tab:dreamerv3_results_sr_ir}
%     \resizebox{\columnwidth}{!}{%
%     \begin{tabular}{c | c c | c c}
%     \toprule
%     %---------------- Header Row 1 ----------------
%     \multirow{2}{*}{\textbf{Train Steps}} 
%     & \multicolumn{2}{c|}{\textbf{Eval Town01} (seen)} 
%     & \multicolumn{2}{c}{\textbf{Eval Town02} (unseen)} \\
    
%     %---------------- Header Row 2 ----------------
%     \cmidrule(lr){2-3} \cmidrule(lr){4-5}
%     & \textbf{SR (\%) \textbf{↑}} & \textbf{IR (\%) \textbf{↓}} 
%     & \textbf{SR (\%) \textbf{↑}} & \textbf{IR (\%) \textbf{↓}} \\
%     \midrule

%     %---------------- Data Row ----------------
%     500K & 64.52 & 35.48 & 64.06 & 35.94 \\

%     \bottomrule
%     \end{tabular}%
%     }
% \end{table}

\subsection{Few-Shots Evaluation}

Table \ref{tab:combined_results} compares three models—InDRiVE (ours), DreamerV3, and DreamerV2—across three driving tasks (Lane Following, Collision Avoidance, and Lane Following + Collision Avoidance) in two CARLA towns, \texttt{Town01} (seen during training) and \texttt{Town02} (unseen). The table reports two primary metrics: Success Rate (SR), the percentage of episodes completed without collisions or lane departures, and Infraction Rate (IR), the percentage of episodes in which a collision or off-lane event occurred. Each model is described by the number of training steps (\emph{Train}) and the number of evaluation steps (\emph{Eval}).

The results highlight several points. First, InDRiVE consistently achieves higher SR and lower IR in both towns, while requiring notably fewer training steps (10K) compared to DreamerV2 or DreamerV3 (510K). In \texttt{Town01}, InDRiVE’s SR ranges from 66\% to 96\% across tasks, while in \texttt{Town02} the performance remains high (83\% to 100\%), indicating strong zero-shot generalization. By contrast, DreamerV2 shows lower SR, particularly in Lane Following tasks, where it struggles to stay within lanes in both towns. DreamerV3 performs moderately well in \texttt{Town01}, and its zero-shot performance in \texttt{Town02} is also decent, but InDRiVE still surpasses it in success rate and infraction reduction. Overall, these findings suggest that incorporating intrinsic disagreement-based exploration (InDRiVE) yields more efficient learning and robust navigation behaviors compared to the Dreamer baselines.

\begin{table*}[!t]
\centering
\caption{Comparison of models on three tasks in both Town01 and Town02}
\label{tab:combined_results}
\renewcommand{\arraystretch}{1.2}
\begin{tabular}{l l c c | c c | c c}
\toprule
\textbf{Task} & \textbf{Model} & \textbf{Train} & \textbf{Eval} &
\multicolumn{2}{c|}{\textbf{Town01} (seen)} &
\multicolumn{2}{c}{\textbf{Town02} (unseen)} \\
\cmidrule(lr){5-6} \cmidrule(lr){7-8}
 &  &  &  & \textbf{SR (\%) \textbf{↑}} & \textbf{IR (\%) \textbf{↓}} &\textbf{SR (\%) \textbf{↑}} & \textbf{IR (\%) \textbf{↓}} \\
\midrule

% ---------------------
% LANE FOLLOWING
% ---------------------
\multirow{3}{*}{LF}
 & \textbf{InDRiVE (ours)} 
    & 10K 
    & 50K 
    & \textbf{96.08}  & \textbf{3.92} 
    & \textbf{100.00} & \textbf{0.00} \\

 & DreamerV3 
    & 510K 
    & 50K 
    & 64.52  & 35.48 
    & 64.06  & 35.94 \\

 & DreamerV2 
    & 510K 
    & 50K 
    & 28.09  & 71.91 
    & 29.07  & 70.93 \\
\midrule

% ---------------------
% COLLISION AVOIDANCE
% ---------------------
\multirow{3}{*}{CA}
 & \textbf{InDRiVE (ours)} 
    & 10K 
    & 50K 
    & 66.10  & 33.90
    & 83.05  & 16.95 \\

 & DreamerV3
    & 510K 
    & 50K 
    & \textbf{73.68}  & \textbf{24.56}
    & \textbf{98.00}  & \textbf{2.00} \\

 & DreamerV2 
    & 510K 
    & 50K 
    & 39.24  & 60.76
    & 33.33  & 66.67 \\
\midrule

% ---------------------
% LANE FOLLOWING + COLLISION AVOIDANCE
% ---------------------
\multirow{3}{*}{LF + CA}
 & \textbf{InDRiVE (ours)} 
    & 10K 
    & 50K 
    & \textbf{83.02}  & \textbf{16.98}
    & \textbf{100.00} & \textbf{0.00} \\

 & DreamerV3
    & 510K 
    & 50K 
    & 73.21  & 26.79
    & 98.00  & 2.00 \\

 & DreamerV2 
    & 510K 
    & 50K 
    & 30.95  & 69.05
    & 28.28  & 71.72 \\
\bottomrule
\end{tabular}
\end{table*}

Fig.~\ref{fig:reward_rate} illustrates three plots compare the average reward rates of InDRiVE (red), DreamerV3 (blue), and DreamerV2 (green) over environment steps in CARLA for three tasks: Lane Following (left), Collision Avoidance (middle), and Lane Following + Collision Avoidance (right). The x‐axis represents the number of environment steps, while the y‐axis denotes the reward rate. Notably, InDRiVE rapidly converges to high reward across all three tasks, whereas the Dreamer baselines require more steps and show greater fluctuation in reward.

\section{Conclusion and Future Work}

We introduced InDRiVE, a fully intrinsic MBRL framework for autonomous driving that eliminates task-specific external rewards by relying solely on ensemble disagreement signals for exploration. Experiments in CARLA show that InDRiVE achieves higher success rates and fewer infractions than DreamerV2 and DreamerV3, while using fewer training steps. Its latent representation transfers effectively to both familiar (\texttt{Town01}) and unfamiliar (\texttt{Town02}) settings, enabling zero-shot or few-shot adaptation to tasks like lane-following and collision avoidance. These findings highlight the benefits of purely intrinsic exploration in uncovering robust driving policies and underscore the potential for reducing dependence on manual reward design. Future research directions include exploring more complex traffic scenarios, integrating richer sensor modalities, addressing sim-to-real transfer, investigating continual and multi-task learning, and evaluating alternative intrinsic reward formulations to further enhance scalability, data efficiency, and adaptability.

%%%%%%%%%%%%%%%%%%%%%%%%%%%%%%%%%%%%%%%%%%%%%%%%%%%%%%%%%%%%%%%%%%%%%%%%%%%%%%%%

% \begin{thebibliography}{99}

% \bibliography{references}

% \bibliographystyle{IEEEtran}
print\bibliography{main}

\begin{thebibliography}{10}
\providecommand{\url}[1]{#1}
\csname url@samestyle\endcsname
\providecommand{\newblock}{\relax}
\providecommand{\bibinfo}[2]{#2}
\providecommand{\BIBentrySTDinterwordspacing}{\spaceskip=0pt\relax}
\providecommand{\BIBentryALTinterwordstretchfactor}{4}
\providecommand{\BIBentryALTinterwordspacing}{\spaceskip=\fontdimen2\font plus
\BIBentryALTinterwordstretchfactor\fontdimen3\font minus \fontdimen4\font\relax}
\providecommand{\BIBforeignlanguage}[2]{{%
\expandafter\ifx\csname l@#1\endcsname\relax
\typeout{** WARNING: IEEEtran.bst: No hyphenation pattern has been}%
\typeout{** loaded for the language `#1'. Using the pattern for}%
\typeout{** the default language instead.}%
\else
\language=\csname l@#1\endcsname
\fi
#2}}
\providecommand{\BIBdecl}{\relax}
\BIBdecl

\bibitem{aubret_information-theoretic_2023}
\BIBentryALTinterwordspacing
A.~Aubret, L.~Matignon, and S.~Hassas, ``An information-theoretic perspective on intrinsic motivation in reinforcement learning: a survey,'' \emph{Entropy}, vol.~25, no.~2, p. 327, Feb. 2023, arXiv:2209.08890 [cs]. [Online]. Available: \url{http://arxiv.org/abs/2209.08890}
\BIBentrySTDinterwordspacing

\bibitem{pathak_curiosity-driven_2017}
\BIBentryALTinterwordspacing
D.~Pathak, P.~Agrawal, A.~A. Efros, and T.~Darrell, ``Curiosity-driven {Exploration} by {Self}-supervised {Prediction},'' May 2017, arXiv:1705.05363 [cs]. [Online]. Available: \url{http://arxiv.org/abs/1705.05363}
\BIBentrySTDinterwordspacing

\bibitem{pathak_self-supervised_2019}
\BIBentryALTinterwordspacing
D.~Pathak, D.~Gandhi, and A.~Gupta, ``Self-{Supervised} {Exploration} via {Disagreement},'' Jun. 2019, arXiv:1906.04161 [cs]. [Online]. Available: \url{http://arxiv.org/abs/1906.04161}
\BIBentrySTDinterwordspacing

\bibitem{burda_exploration_2018}
\BIBentryALTinterwordspacing
Y.~Burda, H.~Edwards, A.~Storkey, and O.~Klimov, ``Exploration by {Random} {Network} {Distillation},'' Oct. 2018, arXiv:1810.12894 [cs]. [Online]. Available: \url{http://arxiv.org/abs/1810.12894}
\BIBentrySTDinterwordspacing

\bibitem{sekar_planning_2020}
\BIBentryALTinterwordspacing
R.~Sekar, O.~Rybkin, K.~Daniilidis, P.~Abbeel, D.~Hafner, and D.~Pathak, ``Planning to {Explore} via {Self}-{Supervised} {World} {Models},'' Jun. 2020, arXiv:2005.05960 [cs]. [Online]. Available: \url{http://arxiv.org/abs/2005.05960}
\BIBentrySTDinterwordspacing

\bibitem{ha_recurrent_2018}
D.~Ha and J.~Schmidhuber, ``Recurrent {World} {Models} {Facilitate} {Policy} {Evolution},'' in \emph{Advances in {Neural} {Information} {Processing} {Systems}}, vol.~31.\hskip 1em plus 0.5em minus 0.4em\relax Curran Associates, Inc., 2018.

\bibitem{gao_dream_2024}
\BIBentryALTinterwordspacing
Y.~Gao, Q.~Zhang, D.-W. Ding, and D.~Zhao, ``Dream to {Drive} {With} {Predictive} {Individual} {World} {Model},'' \emph{IEEE Transactions on Intelligent Vehicles}, pp. 1--16, 2024, conference Name: IEEE Transactions on Intelligent Vehicles. [Online]. Available: \url{https://ieeexplore.ieee.org/document/10547289}
\BIBentrySTDinterwordspacing

\bibitem{hu_model-based_2022}
A.~Hu, G.~Corrado, N.~Griffiths, Z.~Murez, C.~Gurau, H.~Yeo, A.~Kendall, R.~Cipolla, and J.~Shotton, ``\BIBforeignlanguage{en}{Model-{Based} {Imitation} {Learning} for {Urban} {Driving}},'' \emph{\BIBforeignlanguage{en}{Advances in Neural Information Processing Systems}}, vol.~35, pp. 20\,703--20\,716, Dec. 2022.

\bibitem{Dosovitskiy17}
A.~Dosovitskiy, G.~Ros, F.~Codevilla, A.~Lopez, and V.~Koltun, ``{CARLA}: {An} open urban driving simulator,'' in \emph{Proceedings of the 1st Annual Conference on Robot Learning}, 2017, pp. 1--16.

\bibitem{kiran_deep_2022}
B.~R. Kiran, I.~Sobh, V.~Talpaert, P.~Mannion, A.~A.~A. Sallab, S.~Yogamani, and P.~Pérez, ``Deep {Reinforcement} {Learning} for {Autonomous} {Driving}: {A} {Survey},'' \emph{IEEE Transactions on Intelligent Transportation Systems}, vol.~23, no.~6, pp. 4909--4926, Jun. 2022, conference Name: IEEE Transactions on Intelligent Transportation Systems.

\bibitem{hafner_dream_2020}
\BIBentryALTinterwordspacing
D.~Hafner, T.~Lillicrap, J.~Ba, and M.~Norouzi, ``Dream to {Control}: {Learning} {Behaviors} by {Latent} {Imagination},'' Mar. 2020, arXiv:1912.01603 [cs]. [Online]. Available: \url{http://arxiv.org/abs/1912.01603}
\BIBentrySTDinterwordspacing

\bibitem{hafner_mastering_2022}
\BIBentryALTinterwordspacing
D.~Hafner, T.~Lillicrap, M.~Norouzi, and J.~Ba, ``Mastering {Atari} with {Discrete} {World} {Models},'' Feb. 2022, arXiv:2010.02193 [cs]. [Online]. Available: \url{http://arxiv.org/abs/2010.02193}
\BIBentrySTDinterwordspacing

\bibitem{khanzada2024analytical}
\BIBentryALTinterwordspacing
F.~K. Khanzada, B.~Kwon, W.~Jeong, Y.~S. Cho, and J.~Kwon, ``Analytical study on region of interest and dataset size of vision-based end-to-end lateral control for off-road autonomy,'' in \emph{ICRA 2024 Workshop on Resilient Off-road Autonomy}, 2024. [Online]. Available: \url{https://openreview.net/forum?id=KaZ40iwHg7}
\BIBentrySTDinterwordspacing

\bibitem{burda_large-scale_2018-1}
\BIBentryALTinterwordspacing
Y.~Burda, H.~Edwards, D.~Pathak, A.~Storkey, T.~Darrell, and A.~A. Efros, ``Large-{Scale} {Study} of {Curiosity}-{Driven} {Learning},'' Aug. 2018, arXiv:1808.04355 [cs]. [Online]. Available: \url{http://arxiv.org/abs/1808.04355}
\BIBentrySTDinterwordspacing

\bibitem{meyer_possibility_1991}
\BIBentryALTinterwordspacing
J.-A. Meyer and S.~W. Wilson, ``A {Possibility} for {Implementing} {Curiosity} and {Boredom} in {Model}-{Building} {Neural} {Controllers},'' in \emph{From {Animals} to {Animats}: {Proceedings} of the {First} {International} {Conference} on {Simulation} of {Adaptive} {Behavior}}.\hskip 1em plus 0.5em minus 0.4em\relax MIT Press, 1991, pp. 222--227, conference Name: From Animals to Animats: Proceedings of the First International Conference on Simulation of Adaptive Behavior. [Online]. Available: \url{https://ieeexplore.ieee.org/document/6294131}
\BIBentrySTDinterwordspacing

\bibitem{stadie_incentivizing_2015}
\BIBentryALTinterwordspacing
B.~C. Stadie, S.~Levine, and P.~Abbeel, ``Incentivizing {Exploration} {In} {Reinforcement} {Learning} {With} {Deep} {Predictive} {Models},'' Nov. 2015, arXiv:1507.00814 [cs]. [Online]. Available: \url{http://arxiv.org/abs/1507.00814}
\BIBentrySTDinterwordspacing

\bibitem{oudeyer_intrinsic_2007}
\BIBentryALTinterwordspacing
P.-Y. Oudeyer, F.~Kaplan, and V.~V. Hafner, ``Intrinsic {Motivation} {Systems} for {Autonomous} {Mental} {Development},'' \emph{IEEE Transactions on Evolutionary Computation}, vol.~11, no.~2, pp. 265--286, Apr. 2007, conference Name: IEEE Transactions on Evolutionary Computation. [Online]. Available: \url{https://ieeexplore.ieee.org/document/4141061}
\BIBentrySTDinterwordspacing

\bibitem{raileanu_ride_2020-1}
\BIBentryALTinterwordspacing
R.~Raileanu and T.~Rocktäschel, ``{RIDE}: {Rewarding} {Impact}-{Driven} {Exploration} for {Procedurally}-{Generated} {Environments},'' Feb. 2020, arXiv:2002.12292 [cs]. [Online]. Available: \url{http://arxiv.org/abs/2002.12292}
\BIBentrySTDinterwordspacing

\bibitem{noauthor_imagine-2-drive_nodate}
\BIBentryALTinterwordspacing
``Imagine-2-{Drive}: {High}-{Fidelity} {World} {Modeling} in {CARLA} for {Autonomous} {Vehicles}.'' [Online]. Available: \url{https://arxiv.org/html/2411.10171}
\BIBentrySTDinterwordspacing

\bibitem{toromanoff_end--end_2020}
M.~Toromanoff, E.~Wirbel, and F.~Moutarde, ``\BIBforeignlanguage{en}{End-to-{End} {Model}-{Free} {Reinforcement} {Learning} for {Urban} {Driving} {Using} {Implicit} {Affordances}},'' in \emph{\BIBforeignlanguage{en}{2020 {IEEE}/{CVF} {Conference} on {Computer} {Vision} and {Pattern} {Recognition} ({CVPR})}}.\hskip 1em plus 0.5em minus 0.4em\relax Seattle, WA, USA: IEEE, Jun. 2020, pp. 7151--7160.

\bibitem{codevilla_exploring_2019}
\BIBentryALTinterwordspacing
F.~Codevilla, E.~Santana, A.~M. López, and A.~Gaidon, ``Exploring the {Limitations} of {Behavior} {Cloning} for {Autonomous} {Driving},'' Apr. 2019, arXiv:1904.08980 [cs]. [Online]. Available: \url{http://arxiv.org/abs/1904.08980}
\BIBentrySTDinterwordspacing

\bibitem{burda_large-scale_2018}
\BIBentryALTinterwordspacing
Y.~Burda, H.~Edwards, D.~Pathak, A.~Storkey, T.~Darrell, and A.~A. Efros, ``Large-{Scale} {Study} of {Curiosity}-{Driven} {Learning},'' Aug. 2018, arXiv:1808.04355 [cs]. [Online]. Available: \url{http://arxiv.org/abs/1808.04355}
\BIBentrySTDinterwordspacing

\bibitem{hafner_mastering_2024}
\BIBentryALTinterwordspacing
D.~Hafner, J.~Pasukonis, J.~Ba, and T.~Lillicrap, ``Mastering {Diverse} {Domains} through {World} {Models},'' Apr. 2024, arXiv:2301.04104 [cs]. [Online]. Available: \url{http://arxiv.org/abs/2301.04104}
\BIBentrySTDinterwordspacing

\bibitem{li_think2drive_2024}
\BIBentryALTinterwordspacing
Q.~Li, X.~Jia, S.~Wang, and J.~Yan, ``{Think2Drive}: {Efficient} {Reinforcement} {Learning} by {Thinking} in {Latent} {World} {Model} for {Quasi}-{Realistic} {Autonomous} {Driving} (in {CARLA}-v2),'' Jul. 2024, arXiv:2402.16720 [cs]. [Online]. Available: \url{http://arxiv.org/abs/2402.16720}
\BIBentrySTDinterwordspacing

\end{thebibliography}

% \end{thebibliography}

\end{document}